\documentclass[10pt,twocolumn,letterpaper]{article}

\usepackage{iccv}
\usepackage{times}
\usepackage{epsfig}
\usepackage{graphicx}
\usepackage{amsmath}
\usepackage{amssymb}
\usepackage{tabularx}
\usepackage{tablefootnote}

\usepackage[breaklinks=true,bookmarks=false]{hyperref}

\iccvfinalcopy 

\def\iccvPaperID{****} 

\ificcvfinal\fi

\begin{document}

\title{Open-Vocabulary Scene Text Recognition via Pseudo-Image Labeling and Margin Loss}

\author{Xuhua Ren\\
MiHOYO\\
Shanghai, China
\and
Hengcan Shi\\
Monash University\\
Melbourne, Australia\\
{\tt\small shihengcan@gmail.com}
\and
Jin Li\\
Shaanxi Normal University\\
Xian, China\\
}

\maketitle
\ificcvfinal\thispagestyle{empty}\fi

\begin{abstract}
Scene text recognition is an important and challenging task in computer vision. However, most prior works focus on recognizing pre-defined words, while there are various out-of-vocabulary (OOV) words in real-world applications.
In this paper, we propose a novel open-vocabulary text recognition framework, Pseudo-OCR, to recognize OOV words. The key challenge in this task is the lack of OOV training data. To solve this problem, we first propose a pseudo label generation module that leverages character detection and image inpainting to produce substantial pseudo OOV training data from real-world images. Unlike previous synthetic data, our pseudo OOV data contains real characters and backgrounds to simulate real-world applications.
Secondly, to reduce noises in pseudo data, we present a semantic checking mechanism to filter semantically meaningful data. 
Thirdly, we introduce a quality-aware margin loss to boost the training with pseudo data. Our loss includes a margin-based part to enhance the classification ability, and a quality-aware part to penalize low-quality samples in both real and pseudo data.
Extensive experiments demonstrate that our approach outperforms the state-of-the-art on eight datasets and achieves the first rank in the ICDAR2022 challenge.
\end{abstract}

\section{Introduction}

Scene text recognition expects to recognize all characters in an image. It serves as a crucial component in various computer vision applications, such as image captioning \cite{li2019visual}, image generation \cite{hinz2020, ramesh2021zero}, multi-modal systems \cite{gao2020multi, zhao2019image} and image retrieval \cite{wang2019camp,vo2019composing}. 

The existing scene text recognition methods \cite{he2018multi, tang2017scene, shi2016end} can be mainly divided into two types: segmentation-based and context-aware. Segmentation-based approaches \cite{he2018multi, tang2017scene} first segment or detect every character in the image, and then separately recognize them. Nevertheless, they ignore the significant semantic correlations among characters, which are crucial for accurate character recognition. Context-aware works \cite{shi2016end} leverage word-level prior knowledge to learn the relationships among characters to improve text recognition, which usually generates more accurate results than segmentation-based methods. However, they heavily rely on pre-defined words. Real-world applications usually involve diverse words and even new words such as `WASSUP', which are hard to be fully covered by the pre-defined dictionary. 

\begin{figure*}[htb]
\centering
\includegraphics[width=1\textwidth]{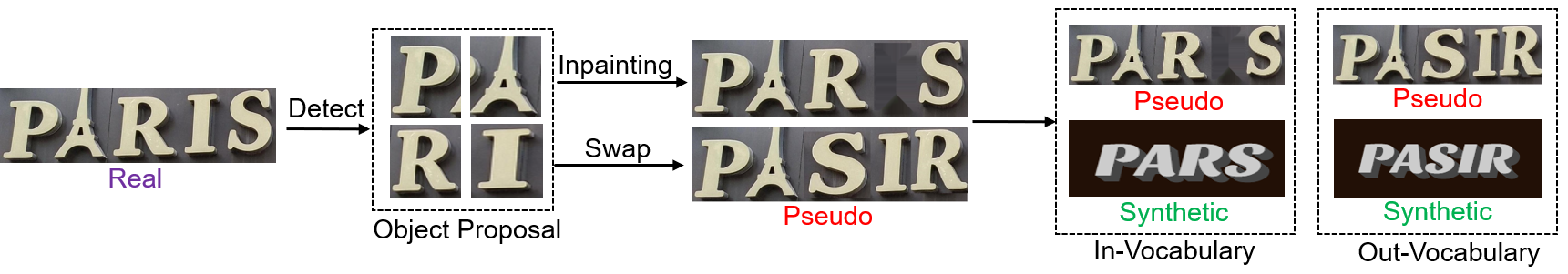}
\caption{Our proposed method builds a pseudo label generation system using real images. For example, the model produces two pseudo labels, `PARS' and `PASIR'  from the real `PARIS' image. The "PARS" is an IV word while "PASIR" is an OOV word. Compared with traditional synthetic images, our pseudo labels are closer to real-world images.}
\label{fig1}
\end{figure*}

Several works \cite{wan2020} notice this challenge and present the open-vocabulary scene text recognition task, which expects to recognize not only in-vocabulary (IV) words but also out-of-vocabulary (OOV) ones. For this task, although prior works \cite{wan2020} employ multiple decoders to improve their generalization ability, the OOV word recognition accuracy is only slightly improved. A key reason is the lack of OOV training samples. To obtain OOV training samples, a straightforward way is data synthesis. Some methods \cite{synthetic, vo2019composing} synthesize pseudo text images to improve text recognition performance. However, as shown in Fig.~\ref{fig1}, these synthetic images only contain computer-generated characters and lack lighting, distortion as well as texture, which are important for scene text recognition in real-world applications.

In this paper, we propose a novel Pseudo-OCR framework for open-vocabulary text recognition, which includes a pseudo label generation module to capture real-image-based pseudo data and a quality-aware margin loss to enhance the model training. Our pseudo label generation module first leverages a character detector to decompose real images into character and background elements, and then generates various OOV pseudo labels from these elements. To obtain more real-like pseudo labels, we employ image inpainting technologies to optimize the appearance of our pseudo labels. In this way, our pseudo labels contain real characters, lighting, distortion and texture, and thus are more suitable for real-world applications, as shown in Fig.~\ref{fig1}. However, there are inevitable noises in pseudo data, such as semantically unreasonable data. Therefore, we propose a semantic checking mechanism to filter and correct our pseudo labels. Specifically, we compare pseudo labels with a large-scale dictionary. If a pseudo label is close to a word in the dictionary, we correct the pseudo label to that word. If a pseudo label largely differs from every word in the dictionary, we filter out it. After our semantic checking, semantically meaningful pseudo labels can be captured.

Meanwhile, a quality-aware margin loss is also presented to better leverage real and our pseudo labels for training. In both pseudo and real training data, low-quality labels (such as unclear characters and missing annotations) affect the model performance. To address challenges, we took inspiration from face recognition, where low-quality faces can significantly reduce performance. Inspired by MagFace \cite{meng2021magface}, we adopted margin loss that improves the classification ability by enlarging class margins. More importantly, we design a quality indicator specific to text recognition images. This indicator is based on the confidence of the character detection model to identify low-quality areas. We then penalize these low-quality samples in the margin loss. 

In summary, our proposed work enhances previous research in three key areas:
\begin{enumerate}
\item We present a novel approach, Pseudo-OCR, to address the open-vocabulary text recognition problem. Pseudo-OCR is able to generate real-image-based pseudo labels for better OOV word recognition. A semantic checking mechanism is also proposed to reduce noises in pseudo labels.
\item We further design a quality-aware margin loss to improve the training with pseudo labels, which contains a margin loss to enhance classification, and a quality indicator to detect and penalize low-quality labels.
\item Experiments show that our approach significantly outperforms previous state-of-the-art methods on eight datasets (i.e., IIIT5k, SVT, IC13, IC13, IC15, IC15, SVTP and CUTE). Moreover, our method achieves the first rank in the ICDAR2022 challenge. 
\end{enumerate}

\section{Related work}

\subsection{Scene Text Recognition}

In the past year, the attention-based decoder \cite{cong2019comparative} has been considered the state-of-the-art pipeline for text recognition, incorporating language modeling, weak character detection supervision, and character recognition in a unified system. Comprised of four essential parts \cite{zhu2016scene}: i) an autoregressively predicting attention-based decoder, ii) a context-modeling Bi-LSTM layer \cite{huang2015b}, iii) a feature-extracting convolutional encoder, and iv) a rectification network to straighten irregular text images, this system distinguishes itself from the Connectionist Temporal Classification (CTC) layer \cite{graves2006} by considering dependencies in the output character space. Several research studies have utilized semantic segmentation for scene text recognition. TextScanner \cite{textscanner}, for example, generates pixel-wise, multi-channel segmentation maps that determine the character class, position, and order. To enhance the recognition process, it also employs an RNN for context modeling. Another study conducted by Liao et al. \cite{liao2019scene} proposed the CA-FCN model, which uses a semantic segmentation network and an attention mechanism for characters. The CA-FCN model also includes a word formation module, enabling it to simultaneously recognize the script and predict the position of each character.


Recently, most context-aware scene text recognition methods use semantics learned from data to enhance recognition \cite{qiao2020seed}. The majority of these approaches utilize Transformers to learn internal language models (LMs) through standard autoregressive (AR) training \cite{parseq}. Unlike previous methods, PARSeq \cite{parseq} utilizes Permuted Language Modeling (PLM) \cite{cui2022pert} instead of standard AR modeling to learn an internal LM. This distinguishes PARSeq from traditional ensemble methods that rely on external LMs for prediction refinement.

However, current methods in scene text recognition do not address the OOV problem, which refers to the difficulties encountered when recognizing words outside of the vocabulary encountered during training. This challenge was first highlighted by Wan et al. \cite{wan2020}, who found that state-of-the-art methods excel at recognizing in-vocabulary words that have been previously seen, but struggle with OOV words. To tackle this issue, the authors introduced a mutual learning strategy that optimizes two types of decoders simultaneously, resulting in improved performance. However, this approach seems to rely solely on ensemble methods to enhance model performance, without considering the issue from the standpoint of the training data.

\subsection{Pseudo Label in Scene Text Recognition}


The self-training approach leverages existing detectors to generate pseudo-labels for unlabeled data. High-confidence pseudo-labels are then selected and used for retraining. Wang et al. \cite{wang2013} employed tracking in videos to acquire difficult examples, which were then used to retrain their detector. This additional data enhanced the detection accuracy of still images. Image synthesis is another method used for generating pseudo-labels. One widely used algorithm for image synthesis is Synthtext\cite{synthetic}. This algorithm utilizes deep learning and segmentation techniques to generate synthetic images of text that blend naturally with existing natural scenes, aligning the text with the geometry of the background image and respecting scene boundaries. Despite its usefulness, Synthtext has some limitations. For instance, it may suffer from semantic incoherence and the presence of non-homogeneous regions. To address these limitations, Zhan et al. proposed the Verisimilar \cite{vo2019composing} method, which leverages the semantic annotations of objects and image regions created through prior semantic segmentation. This method also uses visual saliency to determine the embedding locations within each semantic sensible region, resulting in semantic coherent synthesis. Despite these advancements, the generated images using both Synthtext and Verisimilar methods still have a noticeable gap with real images.

\subsection{Margin Based Loss}

The margin-based softmax loss function is commonly used to train face recognition (FR) models. The addition of margin to the softmax loss improves the discriminative power of the learned features. There are various forms of margin functions introduced in models such as CosFace \cite{wang2018cosface}, and ArcFace \cite{deng2019arcface}. ArcFace is often referred to as an angular margin, while CosFace is known as an additive margin. MagFace \cite{meng2021magface} is a unique approach that assigns different margins based on recognizability, using larger angular margins for high-norm features as they are easier to recognize. Some models \cite{kim2022adaface} consider the margin as a function of image quality, since it can greatly affect which samples provide the strongest gradient or learning signal during training. However, none of these models take into account the unique challenges posed by scene text images, which require a more tailored measure of text image quality, rather than relying solely on image gradient.

\begin{figure*}[htb]
\centering
\includegraphics[width=1.0\textwidth]{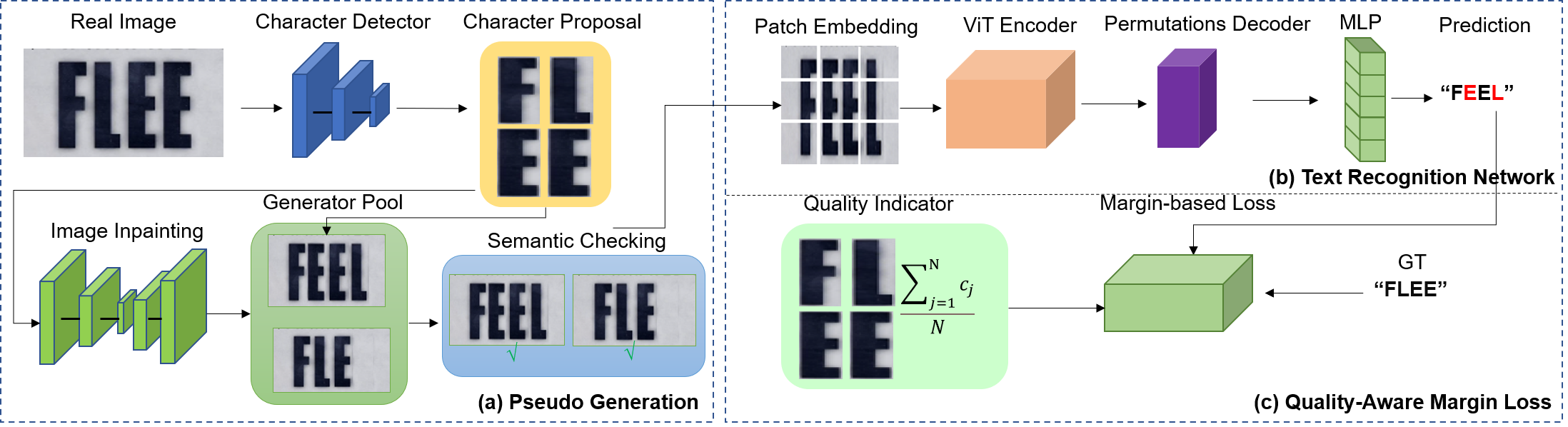}
\caption{
The proposed Pseudo-OCR contains three parts: (a) A pseudo label generation module based on character detector, image inpainting and semantic checking to obtain pseudo labels; (b) a text recognition network with a ViT encoder and a permutation decoder to predict the word in the image; and (c) a quality-aware margin loss including a quality indicator to train the model. In the inference stage, only the text recognition network is used for predicting.}
\label{fig2}
\end{figure*}

\section{Methodology}

\subsection{Overview}

We propose a novel Pseudo-OCR framework that incorporates pseudo-image labeling and margin loss to address the open-vocabulary scene text recognition problem.
Our framework consists of three parts, as illustrated in Fig.~\ref{fig2}. (a) A pseudo label generation model including character detector, image inpainting and semantic checking is first proposed to generate pseudo OOV training samples. (b) A text recognition network recognizes all characters from the input image.
(c) A quality-aware margin loss is used to train the network, and leverages a quality indicator to penalize low-quality samples.  Our proposed pipeline can be trained end-to-end, without the need for any additional manual labeling for the aforementioned modules. During inference, only the recognition network is required, and the pseudo label generation module as well as the loss can be omitted.
Next, we introduce the details of each part.

\subsection{Pseudo label generation}

The first step in pseudo-labeling for scene text recognition is to locate each character. Then, we modify both the image and text to generate new paired data with valid semantics. This system involves four key components: character detector, image inpainting, image augmentation, and semantic checking. The character detector is used to identify and locate each character in the image. Image inpainting is used to fill in the chosen regions of the image. Image augmentation is used to generate a variety of new images from the original image. Finally, semantic checking is used to verify if the generated image-text pair is valid.

 \textbf{Character Detector}: We have used a general object detector, YOLO model, for character detection. In the detection branch of this model, character-level features can be easily obtained by detecting characters within the proposals. To extract both global and character-level features $I$, we applied RoIAlign \cite{he2017mask} to the text proposals generated by FPN. The RoIAlign layer then extracted the features $F(I)$ from the proposals and performed character detection. The extracted features are $7 \times 7$ in size and are fused by element-wise summation. They are then passed through a $3 \times 3$ convolution layer and a $1 \times 1$ layer to create the final fused features for classification $L_{cls}$ and bounding box regression $L_{box}$. The training data for this module is generated synthetically through Synthtext \cite{gupta2016synthetic}, without the use of manual-label data.

 \textbf{Image Inpainting}: We use a UNet-like architecture based on the one presented by Liu et al. \cite{liu2018image}. which incorporates skip connections between mirrored layers, to great effect in object segmentation and image-to-image translation tasks. We apply this mechanism to the up-sampling process, where previous encoding feature maps of the same size are concatenated to retain richer texture and restore the lost background information from the downsampling process. Our training data is generated from Synthtext, where random font, color, and deformation parameters are selected to create styled text which is then rendered onto background images. This generates the corresponding background, foreground text, and overall text all at once.

  \textbf{Image Augmentation}: As depicted in Fig.~\ref{fig2}, we employ two strategies to fabricate novel text image datasets. The initial strategy involves an image inpainting module. We randomly select a character and, cognizant of its specific coordinates, erase it from the image. Subsequently, we feed the entire image into the image inpainting module, resulting in the production of a new image without the selected character. Simultaneously, we also remove the corresponding character from the ground truth, thereby obtaining the updated ground truth for the character. During this process, we eliminate characters with low confidence scores and images with confidence scores that are insufficiently low will not be subjected to the inpainting module. Our second approach involves randomly selecting two characters, harvesting their local images, and swapping their positions. Since the characters come from the same image, it generally does not cause any variations in the image style. By these two strategies, we can generate a multitude of novel datasets for the training of our text recognition models.

\textbf{Semantic Checking}: If newly generated words are meaningless, they can have a detrimental effect on the training process of a model. To address this, we have compiled a comprehensive vocabulary list that includes common words from scene text image recognition datasets, as well as external natural language vocabularies \footnote{\url{https://www.wordfrequency.info/}}. Words that are not included in the list after undergoing random character deletion or swapping are discarded. Additionally, we have implemented classical data augmentation techniques, such as removing the first and last characters, to simulate the truncation effect that often occurs in images.

\subsection{Text recognition network}

We utilize PARSeq \cite{parseq} as our text recognition network, which employs an ensemble of internal AR models with shared weights through PLM. PARSeq follows an encoder-decoder architecture, which is commonly used in sequence modeling tasks. The encoder has 12 layers, while the decoder consists of a single layer.

The ViT is a direct extension of the Transformer architecture to images. A ViT layer contains a single Multi-Head Attention (MHA) module used for self-attention. The encoder is a 12-layer ViT, excluding the classification head. Given an image $x \in \mathbb{R}^{W \times H \times C}$, where $W$ represents the width, $H$ represents the height, and $C$ represents the number of channels, the image is tokenized by dividing it into $t_w \times t_h$ patches of equal size, flattening each patch, and linearly projecting them into dmodel-dimensional tokens using a patch embedding matrix $W_p$. This results in $(W \times H)/(t_{w} \times t_{h})$ tokens. Prior to being processed by the first ViT layer, learned position embeddings of equal dimension are added to the tokens.

The decoder architecture of our model is based on the preLayerNorm Transformer decoder, but it features twice the number of attention heads. The decoder requires three inputs: position tokens, context tokens, and image tokens, along with an optional attention mask. The position tokens encode the target position to be predicted, and each token corresponds directly to a specific position in the output sequence. This parameterization is similar to the query stream of two-stream attention. The attention mask varies depending on how the model is used. During training, we generate masks using random permutations. At inference, the mask can be a standard left-to-right lookahead mask for AR decoding, a cloze mask for iterative refinement, or no mask at all for non-autoregressive (NAR) decoding.

\subsection{Quality-aware margin loss}

In this section, a novel loss function is proposed to learn a more discriminative word recognition model, where the score of character detection is introduced as the penalty term in the MagFace loss. 
In details, the set of image (patch) features in the is indicated as $X \in \mathcal{R}^{d\times N}$, containing $N$ $D$-dimensional features, $x_1, x_2, ..., x_i, ... x_N$. The character-level classifier identifies each feature into $M=94$ classes including letters and common symbols. 

The MagFace loss function is designed to maximize the separation between classes in the feature space, making it an effective solution for open-vocabulary tasks. It ignores irrelevant factors such as background and text in images by focusing on maximizing the cosine similarity between the embeddings of the text images. This results in the model producing distinct and discriminative embeddings for each class, reducing the risk of misclassification and improving the overall performance of the task.

We have designed a loss function that overcomes the limitation of MagFace, which does not consider the quality of scene-text images. This can lead to decreased performance if the data includes a high number of low-quality images. Our approach incorporates a quality indicator module that uses a character detector to provide a more reliable indicator compared to MagFace's image-based magnitude as the quality evaluation indicator. Experimental results show that our method outperforms MagFace and is suitable for scene text images,

\begin{equation}
\begin{split}
    &L_{m}= -\frac{1}{N} \\
    &\sum_{i=1}^N \left(\log \frac{e^{s(\cos(\theta_{y_i} + m(a_i))}}{e^{s(\cos(\theta_{y_i} + m(a_i))} + \sum_{j \neq y_i} e^{s \cos(\theta_j)}} \right).
\label{eq2}
\end{split}
\end{equation}

In Eq. (\ref{eq2}), the term $m(a_i)$ is designed as a monotonically increasing function of the character box confidence $a_i$. Our observations suggest that the confidence of character detection is strongly correlated with the quality of the text image. When the text in the image is clear, the detector can accurately detect the position of the characters. Conversely, when the image quality is poor, the detector often produces low confidence results or fails to detect text boxes altogether. Building on this observation, we have defined our function to account for the varying levels of confidence in character detection that correspond to the quality of the text image,

\begin{equation}
m(a_i) = \frac{u_m - l_m}{u_a - l_a} \times (a_i - l_a) + l_m,
\label{eq3}
\end{equation}

\noindent where $u_m$ and $l_m$ are the upper and lower bounds of the margin $m$, and $u_a$ and $l_a$ are the upper and lower bounds of the input confidence $a_i$. The variable $a_i$ represents the average confidence of $N$ bounding boxes in $i$ -th image and is defined as follows,

\begin{equation}
a_i = \frac{\sum_{j=1}^{N} c_j}{N},
\label{eq4}
\end{equation}

\noindent where $c_j$ is the confidence of the $j$-th bounding box and $N$ is the total number of bounding boxes, our loss introduces an adaptive mechanism to learn a well-structured within-class feature distribution. This is achieved by pulling easy samples towards class centers while pushing hard samples away, which prevents models from overfitting to noisy, low-quality samples, thereby improving scene text recognition performance.

\section{Experiment}
\subsection{Experimental Settings}

\textbf{Training datasets.} Our experimental setup is consistent with previous work in the field. For synthetic training datasets, we utilize MJSynth (MJ) \cite{synthetic} and SynthText (ST) to train proposed modules or compare with other methods. Additionally, we use real data for training, including COCO-Text (COCO) \cite{veit2016coco}, RCTW17 \cite{icdar2017}, Uber-Text (Uber) \cite{zhang2017uber}, ArT \cite{chng2019icdar2019}, LSVT \cite{sun2019icdar}, MLT19 \cite{nayef2019icdar2019}, and ReCTS \cite{zhang2019icdar}. We also leverage two large-scale real datasets based on Open Images: TextOCR \cite{singh2021textocr} and OpenVINO \cite{krylov2021open}.

\textbf{Testing datasets.} We use the IIIT5K \cite{mishra2012scene}, CUTE80 \cite{risnumawan2014robust}, Street View Text (SVT) \cite{wang2011end}, SVT-Perspective (SVTP) \cite{phan2013recognizing}, ICDAR 2013 (IC13) \cite{karatzas2013icdar}, and ICDAR 2015 (IC15) \cite{karatzas2015icdar} datasets for evaluation, following the previous method setting. We use the case-sensitive annotations of Long and Yao \cite{lunrealtext} for IIIT5k, CUTE, SVT, and SVTP. For IC13 and IC15, there are two versions of their respective test splits commonly used in the literature: 857 and 1,015 for IC13; 1,811 and 2,077 for IC15. To avoid confusion, we refer to the benchmark as the union of IIIT5k, CUTE, SVT, SVTP, IC13 (857), IC13 (1,015), IC15 (1,811) and IC15 (2,077).

\textbf{Evaluation metrics.} We use Correctly Recognized Words (CRW) as our metric, which is the percentage of correctly predicted words in the entire dataset. A prediction is considered as correct only if all characters in the recognized text match the corresponding characters in the ground truth word. We reported mean values obtained from three replicates per model. 

\textbf{Character sets}. Varying during both training and inference. Specifically, we use two charsets: a 36-character set containing lowercase alphanumeric characters, and a 94-character set containing mixed-case alphanumeric characters with punctuation.

\begin{table*}
\centering
\begin{tabularx}{1.0\textwidth}{Xccccccccc}
\hline
Method & Train data      & IIIT5k      & SVT      & IC13      & IC13      & IC15      & IC15      & SVTP      & CUTE      \\ \hline
Num     & -       & 3,000        & 647        & 857      & 1,015        & 1,811      & 2,077      & 645      & 288      \\ \hline
\multicolumn{10}{c}{36-char results}                                   \\ \hline
PARSeq \cite{parseq}  & R      & 99.1      & 97.9      & 98.3      & 98.4      & 90.7      & 89.6      & 95.7      & 98.3      \\
S-GTR \cite{sgtr}  & B      & 97.5      & 95.8      & 97.8       & -      & -      & 87.3      & 90.6      & 94.7      \\
CDistNet \cite{cdistnet}  & B      & 96.4      & 93.5      & 97.4      & -      & 86.0      & -      & 88.7      & 93.4      \\
TextScanner \cite{wan2020}  & S      & 95.7      & 92.7      & -      & 94.9      & –      & 83.5      & 84.8      & 91.6      \\
AutoSTR \cite{autostr}  & S      & 94.7      & 90.9      & –      & 94.2      & 81.8      & –      & 81.7      & –      \\
RCEED \cite{cui2021}  & B      & 94.9      & 91.8      & –      & –      & –      & 82.2      & 83.6      & 91.7      \\
PREN2D \cite{yan2021primitive}  & S      & 95.6      & 94.0      & 96.4      & –      & 83.0      & –      & 87.6      & 91.7      \\
STN-CSTR \cite{cai2021}  & S      & 94.2      & 92.3      & 96.3      & 94.1      & 86.1      & 82.0      & 86.2      & -      \\
ViTSTR-B \cite{atienza2021}  & S      & 88.4      & 87.7      & 93.2      & 92.4      & 78.5      & 72.6      & 81.8      & 81.3      \\
CRNN \cite{shi2016end}  & S      & 84.3      & 78.9      & -      & 88.8      & –      & 61.5      & 64.8      & 61.3      \\
TRBA \cite{baek2019wrong}  & S      & 92.1      & 88.9      & -      & 93.1      & –      & 74.7      & 79.5      & 78.2      \\
VisionLAN \cite{wang2021two}  & S      & 95.8      & 91.7      & 95.7      & –      & 83.7      & –      & 86.0      & 88.5      \\ 
\hline
Ours  & R      & 99.4      & 98.2      & 98.6      & 98.5      & 90.9      & 89.8      & 96.1      & 98.2                               \\ \hline \hline
\multicolumn{10}{c}{94-char results for IV words}                                   \\ \hline
OVR  \cite{wan2020}  & R      & 95.19      & 95.45      & 96.23      & 95.78      & 87.12      & 86.78      & 92.77      & 94.13      \\
TRBA \cite{baek2019wrong}  & R      & 96.40      & 96.04     & 95.30      & 95.05      & 87.47      & 86.78      & 93.28      & 95.32      \\
Parseq \cite{parseq} & R      & \textbf{97.36}      & \textbf{97.13}      & 96.77      & 97.31      & 90.0      & 89.35      & \textbf{95.65}      & \textbf{96.54}      \\
Ours  & R      & 97.24     & 96.96      & \textbf{97.52}      & \textbf{97.69}     & \textbf{90.66}      & \textbf{89.95}      & 95.11      & 95.47      \\ \hline
\multicolumn{10}{c}{94-char results for OV words}                                   \\ 
\hline
OVR  \cite{wan2020}  & R      & 89.43      & 79.01      & 88.01      & 88.12      & 55.34      & 57.20      & 80.03      & 86.29      \\
TRBA  \cite{baek2019wrong}  & R      & 89.21      & 78.73      & 88.07      & 88.00      & 55.15      & 57.45      & 79.78      & 86.02      \\
Parseq \cite{parseq}   & R      & 90.43      & 79.79      & 89.66      & 89.60      & 57.74      & 60.0      & 82.01      & 88.25      \\
Ours & R & \textbf{93.91} & \textbf{81.05} & \textbf{92.20} & \textbf{92.20} & \textbf{59.80} & \textbf{62.06} & \textbf{83.27} & \textbf{89.52} \\
\hline
\multicolumn{10}{c}{94-char results for all words}                                   \\ 
\hline
OVR  \cite{wan2020}  & R      & 92.31      & 87.23      & 92.12      & 91.95      & 71.23      & 71.99      & 86.40      & 90.21      \\
TRBA  \cite{baek2019wrong}  & R      & 92.81      & 87.39      & 91.69      & 91.53      & 71.31      & 72.12      & 86.53      & 90.67      \\
Parseq \cite{parseq}  & R      & 93.90      & 88.46      & 93.22      & 93.46      & 73.87      & 74.68     & 88.83      & 92.40      \\
Ours & R & \textbf{95.57} & \textbf{89.00} & \textbf{94.86} & \textbf{94.94} & \textbf{75.23} & \textbf{76.00} & \textbf{89.19} & \textbf{92.49} \\
\hline
\end{tabularx}
\caption{Scene text recognition results, containing 36- and 94-char results. Specifically, we tested our approach on three types of datasets: Synthetic (S) datasets, including MJ and ST; Benchmark (B) datasets, including SVT, IIIT5k, IC13, and IC15; and Real (R) datasets, including COCO, RCTW17, Uber, ArT, LSVT, MLT19, ReCTS, TextOCR, and OpenVINO. Our proposed method outperformed other state-of-the-art methods in both 36- and 94-char experiments, and archived significant improvements for OOV words.}
\label{tab3}
\end{table*}

\subsection{Implementation Details}

Our pseudo label generation module employs YOLOv5s \cite{yolov5} as the character detector. 
The settings of the character detector follow the official config. For image inpainting, we use three image datasets for pretraining: ImageNet, Places2, and CelebA-HQ, and utilize the same settings to train the partial convolutions inpainting model. For the semantic checking module, we not only utilize the training and validation sets in text recognition datasets, but adopt the COCA \footnote{\url{https://www.english-corpora.org/coca/}}, word frequency\footnote{\url{http://www.writewords.org.uk/word_count.asp}}, and iWeb datasets\footnote{\url{https://searchworks.stanford.edu/view/13223153}} as checking lists.

We use ResNet45 \cite{he2016deep} as the backbone in our text recognition network, with weights initialized by the COCO pretrained model. 
We set the input image size to $256 \times 64$ and employ data augmentation techniques including random rotation, image truncation, color jittering, perspective distortion, RandAugment operations (excluding Sharpness), Invert, GaussianBlur, and PoissonNoise. We conducted experiments on 4 NVIDIA V100 GPUs with a batch size of $256$, and trained the network end-to-end using the Adam optimizer with a learning rate of $1\mathrm{e}{-4}$, following the same settings as Parseq \cite{parseq}. 

For our proposed loss, the learning rate is initialized at $0.1$ and decreased by a factor of $10$ at epochs $10$, $20$, and $25$. The training is stopped at the $40$-th epoch. The weight decay and momentum are set to $5\mathrm{e}{-4}$ and $0.9$, respectively. We fix the upper and lower bounds of the text quality as $l_a$ = 0.5, $u_a$ = 1, $l_m$ = 0, and $u_m$ = 6. Finally, our other hyperparameters are also consistent with MagFace \cite{meng2021magface}.

\begin{figure*}
\centering
\includegraphics[width=1.0\textwidth]{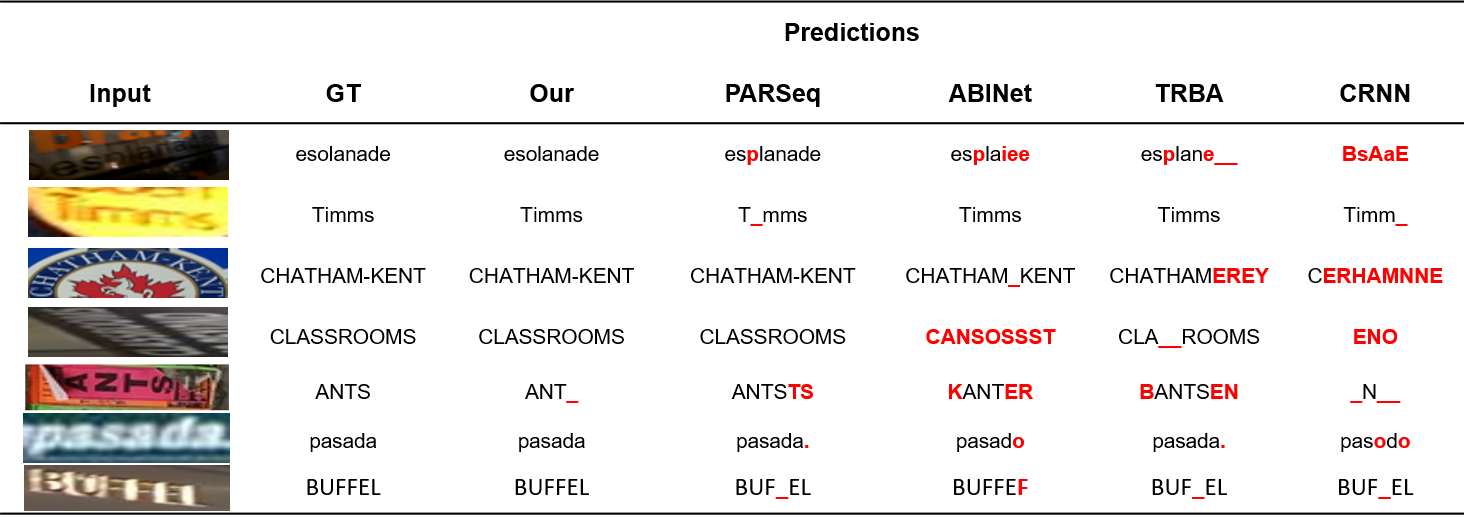}
\caption{Qualitative results for samples taken from various test datasets related to the OOV problem. Both context-free methods, TRBA \cite{baek2019wrong} and CRNN \cite{shi2016end}, were unable to accurately predict certain cases, possibly due to the ambiguity involved. ABINet \cite{fang2022abinet++} encountered difficulties recognizing vertically-oriented and rotated text. PARSeq \cite{parseq} also mis-recognized many characters. Compared with them, our method achieves the best performance.}
\label{fig3}
\end{figure*}

\begin{table}
\centering
\begin{tabularx}{0.5\textwidth}{Xccc}
\hline
                                            & IV: \%        & OOV: \%       & AVE: \%       \\ \hline
\multicolumn{4}{c}{Ablation study for semantic checking module}                                   \\ \hline
w/                             & \textbf{95.06}          & \textbf{81.12}           & \textbf{88.09}             \\
w/o                             & 94.61          & 79.22          & 86.92             \\ \hline
\multicolumn{4}{c}{Ablation study for data augment strategy}                                   \\ \hline
Both                             & \textbf{95.06}          & \textbf{81.12}           & \textbf{88.09}             \\
w/o Remove                               & 95.05          & 80.27          & 87.66             \\
w/o Swap                               & 94.98          & 80.54          & 87.76             \\ \hline
\multicolumn{4}{c}{Compare with other pseudo label method}                                                   \\ \hline
SynthText \cite{gupta2016synthetic}                 & 95.04          & 79.79          & 87.41             \\
UnrealText \cite{lunrealtext}             & 95.03          & 79.70          & 87.37             \\
Edit \cite{wu2019editing}            & 95.03          & 79.65          & 87.34             \\
Ours              & \textbf{95.06}          & \textbf{81.12}           & \textbf{88.09} \\ \hline
\end{tabularx}
\caption{The effects of main components in our pseudo label generation module, and the comparison with other pseudo label generation methods. `W/o' and `w/' mean `without' and `with', respectively. All results are the average of the eight test datasets under the CRW metric.}
\label{tab1}
\end{table}

\subsection{Main Results}

In order to evaluate the effectiveness of our proposed method, we conducted a comparative analysis with popular and recent state-of-the-art methods. To ensure a fair comparison, we not only evaluated our method against published results, but also reproduced state-of-the-art methods. We reported 36- and 94-char results in Table \ref{tab3}. In both 36- and 94-char evaluations, our method outperformed previous methods. In particular, in the more difficult 94-char experiment, our method showed more significant improvements. 

IV and OV results were also shown in Table \ref{tab3}. It could be seen that although previous methods achieved good performance for IV words, they did not recognize OOV words well. In contrast, our proposed Pseudo-OCR yielded remarkable improvements for OOV words. We outperformed our baseline PARSeq \cite{parseq} by up to 3.48\% for OOV words. OVR \cite{wan2020} was also designed for open-vocabulary scene text recognition. However, it only leveraged multiple decoders to improve the generalization ability, while our method generated pseudo training samples for OOV words and achieved improvements of up to 4.48\%. These superior results demonstrated the effectiveness of our pseudo labels and quality-aware margin loss.


\begin{table}
\centering
\begin{tabularx}{0.5\textwidth}{Xccc}
\hline
                                            & IV: \%        & OOV: \%       & AVE: \%       \\ \hline
\multicolumn{4}{c}{Ablation study for margin-based loss}                                   \\ \hline
w/                             & 95.01        & \textbf{80.26}          & \textbf{87.64}             \\
w/o                             & \textbf{95.03}          & 79.68          & 87.36             \\ \hline
\multicolumn{4}{c}{Ablation study for quality indicator}                                   \\ \hline
w/o                              & 95.01          & 80.26          & 87.64             \\
w/ image                               & 95.03          & 80.33          & 87.68             \\
w/ box                              & \textbf{95.03}          & \textbf{80.38}          & \textbf{87.71}             \\ \hline
\multicolumn{4}{c}{Compare with other margin-based loss}                                                   \\ \hline
Arcface \cite{deng2019arcface}                 & 95.01        & 80.26          & 87.64            \\
Megaface \cite{meng2021magface}             & 95.03          & 80.33          & 87.68             \\
Ours              & \textbf{95.03}          & \textbf{80.38}          & \textbf{87.71}             \\ \hline
\end{tabularx}
\caption{The effects of each key component in our quality-aware margin loss, and the comparison with other margin-based losses. `W/o' and `w/' mean `without' and `with', respectively. All results are the average of the eight test datasets under the CRW metric.}
\label{tab2}
\end{table}

\subsection{Ablation Study}

\textbf{The effects of pseudo label generation.} Table \ref{tab1} shows the results of our experiments in terms of IV of CRW score, OOV of CRW score, and their average scores. We conducted experiments to evaluate the performance of our proposed pseudo label generation module, comparing the results using several different settings. We employed semantic checking ("w/") and Remove/Swap in data augmentation ("Both") during implementation, which resulted in our method achieving scores of $95.06\%$, $81.12\%$, and $88.09\%$ for IV, OOV, and their average, respectively. All of which were higher than those achieved with other options.

\textbf{Compare with other pseudo label methods.} Table \ref{tab1} provides a comparative analysis of our proposed method with other state-of-the-art approaches for the same task. Our approach surpasses models that are pre-trained with SynthtText \cite{synthetic}, UnrealText \cite{lunrealtext}, or Edit \cite{wu2019editing}, and subsequently trained on real data, thanks to the integration of our novel pseudo-label module, which utilizes semantic checking and Remove/Swap techniques. Our method's performance metrics are impressive. In particular, our approach leverages semantic knowledge to validate the training data, and correct errors to improve accuracy.

\textbf{The effects of quality-aware margin loss.} Table \ref{tab2} presents the outcomes of our experiments, showcasing the IV and OOV of the CRW score, and their average scores. Our primary aim was to evaluate the effectiveness of our proposed quality-aware margin-based module through various settings. During implementation, we utilized margin-based ("w/") and quality indicator with detector box ("w/ box"), which yielded remarkable results. Our method achieved a score of $95.03\%$, $80.38\%$, and $87.71\%$ for IV, OOV, and their average, respectively. These scores outperformed other alternatives. Overall, our study highlights the potential of the quality-aware margin-based module for enhancing performance in the CRW score evaluation.

\textbf{Compare with other margin-based loss.} Table \ref{tab2} presents a comprehensive comparison of our loss with other margin-based losses. We have outperformed previous models that have utilized cos-like loss, including Arcface \cite{deng2019arcface} and Megaface \cite{meng2021magface}, by incorporating our novel margin-based and quality-aware components. Our method has achieved remarkable performance. Our approach has the added advantage of using the detector confidence score as a quality indicator, which may further enhance the loss convergence. Overall, our study provides valuable insights into the potential of margin-based and quality-aware techniques for enhancing performance in loss functions.

\textbf{Qualitative analysis.} Fig.~\ref{fig3} shows the qualitative results on all test datasets, with the input images displayed in their original orientations and aspect ratios. Incorrect characters are highlighted in red, while missing characters are indicated by a red underscore for predictions that are roughly aligned with the ground truth. Both context-free methods, TRBA \cite{baek2019wrong} and CRNN \cite{shi2016end}, did not accurately predict certain cases, possibly due to the ambiguity involved. ABINet \cite{fang2022abinet++} encountered difficulties in recognizing vertically-oriented images. Considering the presence of obstructions on the image, such as occlusion, we believe that the recognition results of all models in the last row of the image are suboptimal and that this cannot be avoided. The results demonstrate that our model can accurately identify OOV images and maintain good performance on IV images.

\subsection{Comparison in ICADR 2022 Challenge}

\begin{table}
\centering
\begin{tabular}{ccccc}
\hline
       & \multicolumn{1}{c|}{AVG: \%} & \multicolumn{1}{c|}{IV: \%} & \multicolumn{1}{c|}{OOV: \%} & 
       \multicolumn{1}{c}{Ensemble} 
       \\ \hline
Method   & CRW      & CRW                & CRW        \\
\hline
Ours    & \textbf{70.98}                       & \textbf{82.81}           & 59.15      \\
OCRFLYV2   & 70.31                      & 81.02           & \textbf{59.61}      & \checkmark\\ 
HuiGuanV2    & 70.28                      & 81.73           & 58.83     &\checkmark\\ 
oov3decode    & 70.22                       & 81.58           & 58.86     &\checkmark\\
ViT-based  & 70.00                        & 81.36             & 58.64    &\checkmark\\
\hline
\end{tabular}
\caption{Comparison of the proposed method with other state-of-the-art methods on the on-site test set of the ICDAR Out-of-Vocabulary Challenge. Our method achieves the first rank.}
\label{tab4}
\end{table}

Over 200 teams participated in the ICDAR Out of Vocabulary challenge, utilizing official training and validation data from various datasets including ICDAR13, ICDAR15, MLT19, COCO-Text, TextOCR, HierText, and OpenImagesText. This challenge focuses on the cropped word recognition task, where participants are required to predict the recognition of all cropped words in the test set. The test set contains 313,751 cropped words, out of which 271,664 are IV words and 42,087 are OOV ones. According to Table \ref{tab4}, our method outperformed other methods in the challenge. Notably, our method achieved this without utilizing the ensemble strategy. 

\section{Conclusion}

In this paper, we have presented a Pseudo-OCR framework for open-vocabulary scene text recognition. Our approach entails two major contributions. Firstly, we propose a novel pseudo label generation module that integrates character detection and image inpainting techniques to create a large volume of training data. Different from previous synthetic labels, our pseudo labels are closer to real-world images. Meanwhile, we design a text semantic checking in our pseudo label generation module to filter semantically meaningful pseudo data. Moreover, we introduce a margin loss that optimizes geodesic distance margins to mitigate the effect of low-quality samples. We also present a text quality indicator to dynamically adjust the margin of each class based on image quality. Extensive experimental results demonstrate that our method outperforms the state-of-the-art on various test datasets, enhancing both IV and OOV performance. Moreover, our approach secured the first position in the ICDAR2022 challenge.

{\small
\bibliographystyle{ieee_fullname}
\bibliography{egbib}
}

\end{document}